%% file: main.tex
\documentclass[letterpaper, 10 pt, conference]{ieeeconf}  

\IEEEoverridecommandlockouts

\usepackage{graphics}
\usepackage{graphicx}
\usepackage{xspace}
\usepackage[table, dvipsnames]{xcolor}
\usepackage[font=small, labelfont=bf]{caption}
\usepackage[colorlinks = true,
            linkcolor = black,
            urlcolor  = blue,
            citecolor = magenta,
            bookmarks = true, pagebackref]{hyperref}
\usepackage{lipsum}
\usepackage{xspace}
\usepackage{todonotes}
\usepackage{amsmath}
\usepackage{amssymb}
\usepackage{booktabs}
\usepackage{array}
\usepackage{soul}
\usepackage{pifont}
\usepackage{float}
\usepackage{arydshln}
\usepackage{flushend}
\usepackage{cleveref}
\usepackage{multirow}
\usepackage{cite}
\usepackage{algpseudocodex}

\newcommand{\mybaseline}[0]{NavStack\xspace}
\newcommand{\mybaselinelong}[0]{ROS NavStack\xspace}
\newcommand{\mymethod}[0]{CANVAS\xspace}
\newcommand{\mymethodlong}[0]{Commonsense-Aware NaVigAtion System\xspace}
\newcommand{\mydata}[0]{COMMAND\xspace}
\newcommand{\mydatalong}[0]{COMMonsense-Aware Navigation Dataset\xspace}

\newcommand{\myparagraph}[1]{\noindent\textbf{#1}}
\newcommand{\ours}[0]{\rowcolor{LimeGreen!30}}

\title{\LARGE \bf \mymethod: Commonsense-Aware Navigation System \\ for 
Intuitive Human-Robot Interaction}

\author{
Suhwan Choi$^{\dagger1}$,
Yongjun Cho$^{\dagger1}$,
Minchan Kim$^{\dagger1}$,
Jaeyoon Jung$^{\dagger1}$,
Myunchul Joe$^{1}$,
Yubeen Park$^{1}$,
\\[0.5em]
Minseo Kim$^{2}$,
Sungwoong Kim$^{2}$,
Sungjae Lee$^{2}$,
Hwiseong Park$^{1}$,
Jiwan Chung$^{2}$,
Youngjae Yu$^{2}$
\\[0.5em]
$^\dagger$Equal Contribution. $^{1}$MAUM.AI, $^{2}$Yonsei University.
\\[0.5em]
\texttt{\{milkclouds, cyjun0304, alan, jyjung\}@maum.ai}
\thanks{
Videos, datasets, and models: \href{https://worv-ai.github.io/canvas}{worv-ai.github.io/canvas}
}
}

\begin{document}

\makeatletter
\let\@oldmaketitle\@maketitle%
\renewcommand{\@maketitle}{\@oldmaketitle%
    \centering
    \vspace*{4mm}
    \includegraphics[width=\textwidth]{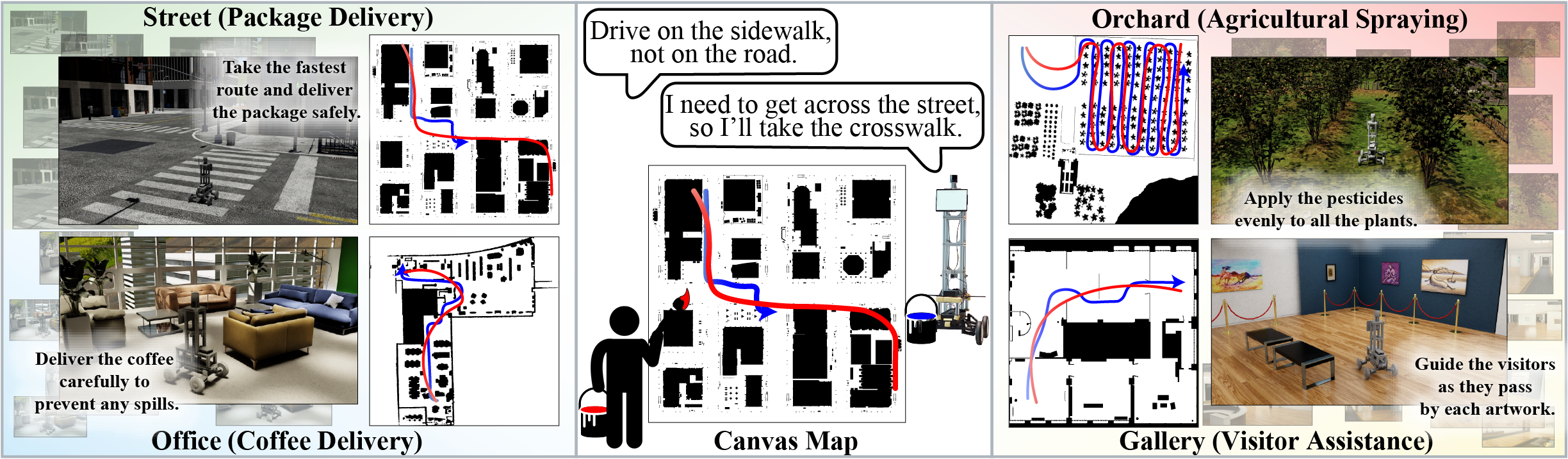}
    \captionof{figure}{Humans often give abstract navigation directions using simple instruction, relying on the recipient's commonsense to bridge the gaps. With \mymethod, robots can interpret and act on these directions like humans do, fostering a shared understanding of the environment. It shows how robots can use commonsense to translate vague human instructions into concrete actions, navigating across diverse settings in our \mydata dataset, which we plan to open-source as valuable resources for imitation learning in commonsense-aware navigation.
}
    \label{fig:1_teaser}
    \vspace*{1mm}
}
\makeatother
\maketitle
\IEEEpeerreviewmaketitle
\setcounter{figure}{1}

\input{sections/0_abstract}
\input{sections/1_introduction}
\input{sections/2_related_work}
\input{sections/3_dataset}
\input{sections/4_method}
\input{sections/5_experiments}
\input{sections/6_conclusion}
\input{sections/acknowledgment}

\hypersetup{linkcolor=Red, urlcolor=Blue}
\bibliographystyle{IEEEtran}
\bibliography{refs}

\end{document}

%% file: sections/0_abstract.tex
\begin{abstract}
Real-life robot navigation involves more than just reaching a destination; it requires optimizing movements while addressing scenario-specific goals. 
An intuitive way for humans to express these goals is through abstract cues like verbal commands or rough sketches. 
Such human guidance may lack details or be noisy. Nonetheless, we expect robots to navigate as intended. 
For robots to interpret and execute these abstract instructions in line with human expectations, they must share a common understanding of basic navigation concepts with humans.
To this end, we introduce \mymethod, a novel framework that combines visual and linguistic instructions for commonsense-aware navigation.
Its success is driven by imitation learning, enabling the robot to learn from human navigation behavior.
We present \mydata, a comprehensive dataset with human-annotated navigation results, spanning over 48 hours and 219 km, designed to train commonsense-aware navigation systems in simulated environments.
Our experiments show that \mymethod outperforms the strong rule-based system \mybaselinelong across all environments, demonstrating superior performance with noisy instructions.
Notably, in the orchard environment, where \mybaselinelong records a 0\% total success rate, \mymethod achieves a total success rate of 67\%.
\mymethod also closely aligns with human demonstrations and commonsense constraints, even in unseen environments.
Furthermore, real-world deployment of CANVAS showcases impressive Sim2Real transfer with a total success rate of 69\%, highlighting the potential of learning from human demonstrations in simulated environments for real-world applications.
\end{abstract}

%% file: sections/1_introduction.tex
\section{Introduction}
\label{sec:1_introduction}

Real-life robot navigation scenarios involve addressing complex objectives that extend far beyond simply reaching a destination. 
For example, an agricultural spraying robot must maximize field coverage~\cite{Li2024orchard}, while a package delivery robot must adhere to road lanes and use crosswalks when transitioning between sidewalks.~\cite{wang2022motion,8793608}
In both cases, robots need to optimize their movements while responding to the specific requirements of the scenario.

Humans typically communicate these scenario-specific goals through high-level guidance, such as verbal commands~\cite{Anderson_2018_CVPR, Huang2023VLMaps, zhang2024navid}, rough sketches of the desired route~\cite{skubic2007using, boniardi2016sketch}, or a combination of both~\cite{lim2n}.
While such guidance outlines the robot’s overall objectives, it often lacks the specificity required for precise execution.
To convert these abstract and imprecise instructions into actionable navigation plans, robots need commonsense knowledge.
In the context of robotics, commonsense refers to the general understanding humans naturally use to make decisions, covering aspects such as human desires, physics, and causality~\cite{toberg2024commonsense}. Robots must leverage this knowledge to flexibly adjust their paths, ensuring their decisions align with human expectations by adhering to commonsense constraints posed by the environment and the user's true intentions.

In response to these challenges, we introduce \mymethod (\mymethodlong), a novel framework for integrating abstract human instructions into robot navigation.
Our approach utilizes both visual and linguistic inputs, such as rough sketch trajectories on map images or textual descriptions.
These multimodal instructions are processed by a vision-language model that generates incremental navigation targets.
By leveraging the commonsense knowledge embedded in pre-trained vision-language models~\cite{chen2020commonsense, zhou2023esc, navgpt, chen2024commonsense, kim2024exploiting}, robots can develop a versatile understanding of commonsense navigation dynamics. 
Quantifying successful navigation behaviors using rewards\cite{DD-PPO, wireless-nav, taheri2024deep} is particularly difficult in commonsense-aware navigation. 
Therefore, we employ imitation learning, enabling the robot to comprehend user intent behind noisy and imprecise instructions from human demonstrations. \cite{goalGAIL}

Additionally, we introduce \mydata (\mydatalong), a comprehensive dataset designed to train commonsense-aware navigation robots.
The dataset features three simulated environments with distinct characteristics (office, street, and orchard).
To facilitate imitation learning for instruction-following tasks, we provide 3,343 fully human-annotated navigation results from the simulated environments. Notably, \mydata offers 48 hours of driving data, which is nearly three times longer than GoStanford~\cite{hirose2019deep}, covering 219 km and thereby enriching the dataset's diversity and scope.
Furthermore, we propose two metrics to evaluate the commonsense adherence of navigation algorithms: Trajectory Deviation Distance (TDD) and Instruction Violation Rate (IVR).

Our results show that \mymethod consistently outperforms \mybaselinelong~\cite{rosnavstack} across all environments with noisy sketch instructions.
Particularly in the challenging orchard environment, \mymethod navigates effectively, while \mybaseline fails due to its reliance on rule-based algorithms\cite{orchard-ifac}. \mymethod’s trajectory closely mirrors human demonstrations with fewer commonsense constraint violations, indicating a better understanding of human expectations. Despite being trained only on simulated data, \mymethod also excels in real-world scenarios, demonstrating strong Sim2Real transfer capabilities.

Our contributions are threefold:

\begin{enumerate}
    \vspace{-0.25em}
    \setlength{\itemsep}{0.1em}
    \item We introduce \mymethod, a novel framework that allows humans to easily communicate with robots using multimodal inputs, ensuring that robots effectively achieve navigation goals, even when human instructions are vague or noisy.
    \item We introduce \mydata, a dataset for training commonsense-aware navigation robots, featuring 48 hours of driving data over 219 kilometers, with fully human-annotated sketch instruction and navigation outcomes.
    \item We present extensive experiments demonstrating that \mymethod outperforms \mybaselinelong in success rate, collision rate, trajectory deviation distance, and instruction violation rate. To support further research, we are open-sourcing CANVAS and COMMAND for imitation learning in commonsense-aware robot navigation.
    \vspace{-0.25em}
\end{enumerate}

%% file: sections/2_related_work.tex
\section{Related Work}
\label{sec:2_related_work}

\begin{figure*}[t]
\centering
\vspace*{2mm}
\includegraphics[width=\textwidth]{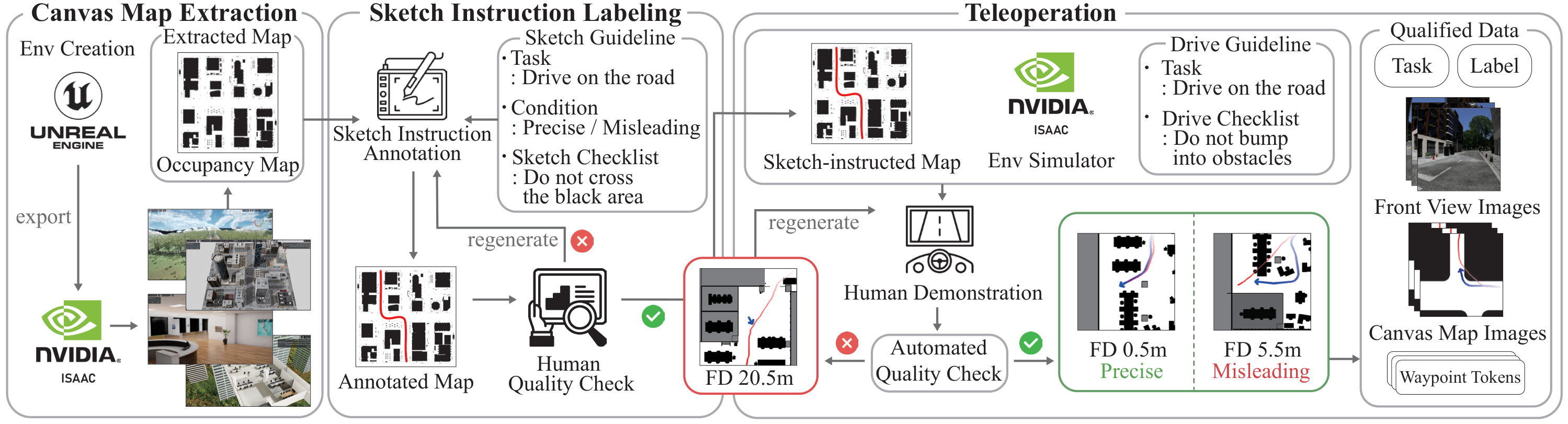}
\caption{Data collection pipeline for \mydata dataset.
(a) First, we create diverse navigation environments and extract maps. (b) Then, human annotators sketch routes on the maps based on the guidelines. (c) Finally, we use teleoperation to collect human demonstrations. \textcolor{red}{Red line} shows the roughly sketched routes, while the \textcolor{blue}{blue line} shows the human-demonstrated trajectory. FD refers to \textit{Frechet distance}.
}
\label{fig:2_data_pipeline}
\vspace*{-4mm}
\end{figure*}

\subsection{Robot Navigation}

Historically, robot navigation systems were largely rule-based~\cite{gul2019comprehensive, orchard-ifac, slam-nav}, relying on a set of predefined rules, as seen in frameworks like the \mybaselinelong~\cite{rosnavstack}.
Following the successful application of deep learning to robotics, more flexible neural navigation approaches have emerged.
Visual navigation models, such as NoMaD~\cite{nomad}, ViNT~\cite{vint}, and GNM~\cite{gnm}, utilize images as goal representations.
However, since their high-level planning heavily relies on the topological graph with first-person visual observations, they cannot handle unvisited locations and are sensitive to environmental changes. \cite{topomap}
Vision-language navigation integrates visual information from sensors with language instructions~\cite{gu-etal-2022-vision, Huang2023VLMaps, zhang2024navid}. 
However, the ambiguous nature of language instructions poses limitations on controlling detailed navigation routes.~\cite{wu2023vision}
Recently, LIM2N~\cite{lim2n} enabled users to control robots through natural language and sketch trajectories, combining high-level goals with precise motion paths for more intuitive interaction. 
However, the system's demand for highly accurate and detailed instructions, coupled with its vulnerability to missing details or small mistakes, reduces its usability for non-expert users and limits its effectiveness in a broader range of real-world applications \cite{Anderson_2018_CVPR}.

Our proposed method, \mymethod, differentiates itself by addressing the challenge of interpreting abstract or noisy human instructions.
\mymethod converts visual and linguistic instructions into detailed navigation actions, utilizing commonsense knowledge to fill in the gaps.
This integration of commonsense enables \mymethod to dynamically adapt its navigation strategies across diverse contexts, resulting in enhanced task execution compared to \mybaselinelong.

\input{tables/1_related_work}

\subsection{Imitation Learning in Robotics}
\label{imitation_learning_in_robotics}

Imitation Learning (IL) enables agents to learn tasks by mimicking expert demonstrations, eliminating the need for predefined rules or reward functions typically required in Reinforcement Learning (RL)~\cite{imitation-zeroshot}.
By directly leveraging expert behavior, IL has proven particularly advantageous in situations where designing a reward function is challenging or exploration involves potential risks~\cite{zare2024survey}.
As a result, there has been a growing interest in applying IL to robot navigation~\cite{voila,vln-zeroshot}.
A key challenge in IL is modeling multimodal action distributions~\cite{shafiullah2022behavior}.
One solution is to quantize actions into discrete tokens, simplifying the action space~\cite{metz2017discrete,dadashi2021continuous,shafiullah2022behavior,chebotar2023q,vq-bet}.
Autoregressive prediction of quantized actions effectively reduces the complexity of modeling diverse and feasible action sequences. 

\mymethod builds upon this idea by converting continuous waypoints into 128 discrete waypoint tokens using K-means clustering~\cite{rt-2, li2024hydra}.
This approach enhances the ability of robots to model multimodal action distributions, enabling robust navigation strategies that adapt to diverse human instructions and environmental variations.

%% file: tables/1_related_work.tex
\begin{table}[h]
    \centering
    \setlength{\tabcolsep}{1.5pt}
    \resizebox{\linewidth}{!}{
        \begin{tabular}{l|c|c|c|c}
            \cline{2-5}
            \multicolumn{1}{c|}{} & \textbf{\textit{NoMaD}~\cite{nomad}} & \textbf{\textit{NaVid}~\cite{zhang2024navid}} & \textbf{\textit{LIM2N}~\cite{lim2n}} & \multicolumn{1}{>{\columncolor{LimeGreen!30}}c}{\textbf{\mymethod}} \\
            \hline
            \textbf{Instruction} & Image & Language & Sketch, Language & Sketch, Language \\
            \textbf{Misleading (\ref{misleading})} & \textcolor{red}{\ding{55}} & \textcolor{red}
            {\ding{55}} & \textcolor{red}{\ding{55}} & \textcolor{green}{\ding{51}} \\
            \textbf{Custom Dataset} & \textcolor{red}{\ding{55}} & \textcolor{green}{\ding{51}} & \textcolor{green}{\ding{51}} & \textcolor{green}{\ding{51}} \\
            \textbf{Environment} & Real & Real & Real, Simulation & Real, Simulation \\
            \textbf{Scenes} & Indoor, Outdoor & Indoor & Indoor & Indoor, Outdoor \\
            \hline
        \end{tabular}%
    }
    \caption{Comparison between various robot navigation methods.}
    \label{tab:1_related_work}
    \vspace*{-4mm}
\end{table}

%% file: sections/3_dataset.tex
\section{Dataset and Task}
\label{sec:3_dataset}

An interactive robot navigation framework should fulfill two key objectives: first, humans should be able to communicate desired routes and requirements intuitively; second, the robot should accurately interpret and execute those instructions.
However, achieving these goals can be challenging.
Simplifying communication for humans often complicates it for robots because humans naturally assume the listener shares their commonsense knowledge.
This commonsense allows humans to infer meaning even when instructions are incomplete or imprecise but also causes robots to struggle without explicit and precise input. 

To address this challenge, we introduce COMMAND—a comprehensive experiment suite designed to assess whether robots can use commonsense understanding to transform noisy or abstract human instructions into the most desired trajectory.
\mydata was collected in three distinct environments: office, street, and orchard, using NVIDIA Isaac Sim~\cite{isaacsim}.
The data consists of high-quality sketch instruction labels and teleoperation data, \textbf{all by human experts}.
\mydata contains 48 hours of driving data covering a distance of 219 kilometers.
In this section, we describe the dataset curation process and its resulting task definition.

\subsection{Dataset}
A datapoint in COMMAND includes a canvas map, sketch and language instructions, commonsense constraints, and teleoperation records.
The canvas maps, linked to each environment, provide occupancy information and serve as a human-robot communication interface.
We assume humans provide instructions in two modalities:
sketch instructions \( S \) that consist of hand-drawn trajectories on maps for the robot to follow, and language instructions \( L \) that outline goals and related requirements.
When collecting sketch instructions \( S \), we introduce both \textbf{\textit{Precise}} and \textbf{\textit{Misleading}} conditions to gather training data that enables our model to handle noisy sketch instructions more robustly.
In addition to the instructions, we define a set of commonsense constraints \( C \), which are derived from the navigation environment \( E \) and the language instructions \( L \).
These constraints help evaluate whether the robot exhibits appropriate navigation behaviors, such as using crosswalks.
We also include human teleoperation records to optimize and evaluate the robot behavior against human actions.
An overview of our data collection pipeline is illustrated in Figure~\ref{fig:2_data_pipeline}, and the statistics are provided in Table~\ref{tab:2_dataset_statistics}.
We detail each step below.

\subsubsection{Environments} COMMAND features three simulated environments, each tailored to specific scenarios.
The simulated environments include tasks such as coffee delivery in an office, package delivery on the street, and agricultural spraying in an orchard.
An expert designer creates each simulated environment using Unreal Engine~\cite{unrealengine}.

\subsubsection{Canvas Map Extraction} The simulated environments are subsequently exported to NVIDIA Isaac Sim, where occupancy maps are extracted programmatically.

\subsubsection{Sketch Instruction Labeling}\label{misleading} The data workers draw sketch trajectories on the canvas maps following sketch guidelines manually crafted by the authors.
When the \textit{Precise} condition is included, the workers trace the most efficient route, closely following the guidelines.
In contrast, when the \textit{Misleading} condition is included, data workers are instructed to deliberately introduce noise by drawing trajectories that pass through walls or objects.
All sketch instructions undergo manual inspection to ensure quality.

\subsubsection{Teleoperation} The data workers are then provided both the sketch and language instructions to teleoperate the virtual robot in NVIDIA Isaac Sim.
We record front view images, canvas map images, and the human-demonstrated trajectory.
After collecting the teleoperation data, we adopt the \textit{Frechet distance} (FD)~\cite{aronov2006frechet} to measure the discrepancy between the sketch trajectory and the human-demonstrated trajectory.
This metric, which indicates noise in the sketch instructions, tends to be higher in the \textit{Precise} condition and lower in the \textit{Misleading} condition. 

\input{tables/2_long_dataset_statistics}

\subsection{Task Definition}

At each timestep \( t \), the robot \( R \) manages two states: the front view image \( X_f(t) \) and the robot's hindsight trajectory~\cite{rt-trajectory} up to timestep \( t-1 \), denoted as \( H(t) \).
The front view image \( X_f(t) \) is captured by the robot’s camera, while \( H(t) \) is tracked through odometry to log the robot’s past positions.
We combine the sketch instruction \( S \) and hindsight trajectory \( H(t) \) onto the same map to create the canvas map image \( X_c(t) \).
At each step, the robot \( R \) generates an action \( y(t) = [w_0, w_1, w_2, w_3] \), which is a sequence of waypoints.
This action is conditioned on the front view image \( X_f(t) \), the canvas map image \( X_c(t) \), and the language instruction \( L \). Formally, the robot's action is defined as:
\[
y(t) = R(X_f(t), X_c(t), L)
\]
At the end of each iteration, the hindsight trajectory is updated by appending \(p(t)\), which represents the robot’s position updated through predicted waypoints \(y(t)\), resulting in \( H(t+1) = (p(1), p(2), ..., p(t)) \).
This update is reflected in the next canvas map image \( X_c(t+1) \), while the front view image \( X_f(t+1) \) is also updated based on the robot's new position.
This process continues until the robot either reaches the destination or a maximum timestep \( t=T \) is reached.

%% file: tables/2_long_dataset_statistics.tex
\begin{table*}[h]
\vspace{2mm}
    \resizebox{\linewidth}{!}{
        \centering
        \begin{tabular}{l|cccc|cccccc}
            \hline
            Split & \multicolumn{4}{c|}{Train} & \multicolumn{6}{c}{Test} \\
            \hline
            \multirow{2}{*}{Environment} & \multirow{2}{*}{Office} & \multicolumn{2}{c}{Street} & \multirow{2}{*}{Orchard} & \multirow{2}{*}{Office} & \multicolumn{2}{c}{Street} & \multirow{2}{*}{Orchard} & \multirow{2}{*}{Gallery} & Real \\
            \cline{3-4}\cline{7-8}
            & & \multicolumn{1}{c}{Road} & \multicolumn{1}{c}{Sidewalk} & & & Road & Sidewalk & & & Office \\
            \hline
            Count & 2,263 & 403 & 410 & 267 & 10 & 20 & 10 & 10 & 10 & 10 \\
            Avg. Time & 31s & 57s & 103s & 172s & 39s & 56s & 127s & 182s & 76s & 30s \\
            Avg. Distance & 32.8m & 80.4m & 150.0m & 191.6m & 38.7m & 91.0m & 152.0m & 232.88m & 48.8m & 16.0m \\
            Avg. FD (P / M) & 1.05 / 1.77 & 0.97 / 2.02  & 3.03 / 3.50 & 1.91 / 3.76 & 0.77 / 1.62 & 1.28 / 1.65 & 1.32 / 2.33 & 1.51 / 2.27 & 0.68 / 1.36 & 1.44 / 1.63\\
            \% of Misleading & 31\% & 51\% & 51\% & 40\% & 50\% & 50\% & 50\% & 50\% & 50\% & 50\%\\
            \hline
        \end{tabular}
    }
    \caption{Statistics for the train and test set. The train dataset includes 48 hours of driving data over 219 kilometers, while the test dataset consists of 1.6 hours. FD refers to \textit{Frechet distance}, where (P / M) stands for Precise and Misleading. The unit of FD is meters.}
    \label{tab:2_dataset_statistics}
    \vspace*{-4mm}
\end{table*}

%% file: sections/4_method.tex
\section{Method}
\label{sec:4_method}

To address the problem formulated in Section~\ref{sec:3_dataset}, we introduce \mymethod, a navigation system designed to bridge the gap between abstract human instructions and concrete robot actions by leveraging commonsense understanding from pre-trained vision-language models (VLMs)~\cite{idefics2}.
In this section, we provide an overview of the model architecture, as well as the training and inference processes.

\subsection{Architecture}

\begin{figure}[t]
\centering
\includegraphics[width=0.45\textwidth]{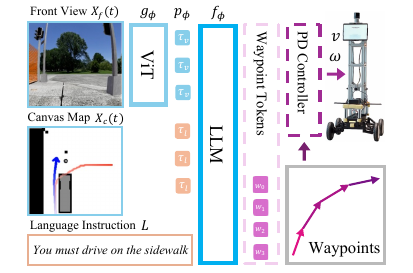}
\caption{Overview of the \mymethod framework. It processes the front view image \(X_f(t)\), canvas map \(X_c(t)\), and language instruction \(L\) to generate waypoint tokens, which are passed to a PD controller to move the robot.}
\label{fig:3_framework}
\vspace{-4mm}
\end{figure}

The model architecture is illustrated in Figure \ref{fig:3_framework}.
We adopt a VLM denoted as \( \pi_{\theta} \).
The front view image \( X_f(t) \) and canvas map image \( X_c(t) \) are processed through a vision encoder \( g_{\phi}(\cdot) \).
This results in two visual features, \( Z_f = g_{\phi}(X_f(t)) \) and \( Z_c = g_{\phi}(X_c(t)) \).
A projector \( p_{\phi} \) is used to project these visual features into the word embedding space, producing a sequence of visual tokens \( \tau_v = p_{\phi}(Z_f, Z_c) \).
A sequence of language tokens \( \tau_l \) is also obtained from the language instruction \( L \).
Both the visual tokens \( \tau_v \) and language tokens \( \tau_l \) are then fed into the large language model denoted as \( f_{\phi}(\cdot) \), which outputs the waypoint tokens \( [w_0, w_1, w_2, w_3] = f_{\phi}(\tau_v, \tau_l) \).
Due to the reasons mentioned in Section \ref{imitation_learning_in_robotics}, we apply the simplest method, K-means clustering, to discretize continuous waypoints into tokens, with empirical testing showing that K=128 outperformed 32, 64, or 256. Fewer tokens can hinder precise actions like navigating narrow passages.

\subsection{Training}

\mymethod is designed to generate actions as a sequence of waypoint tokens.
A robot's trajectory can be represented by \( N \) consecutive waypoints.
During training, we minimize the negative log-likelihood loss, which is formulated as follows:
\[
J(\pi_{\theta}) = \sum_{n=1}^{N} \sum_{t=0}^{3} \log \pi_{\theta}\left(w_t^n \mid X_f(t)^n, X_c(t)^n, L^n\right)
\]

The model reframes the navigation as a classification problem, where it predicts the next waypoint based on the current state and given instructions.
As explained in Related Work Section~\ref{imitation_learning_in_robotics}, by treating navigation as a classification task, the model can manage multimodal distributions, enhancing both stability and accuracy in complex environments.

\subsection{Inference}

During inference, the model-generated discrete waypoint tokens convert into continuous waypoints, which are then input into a Proportional-Derivative (PD) controller to produce linear and angular velocities \( (v, \omega) \) for the robot’s actuators.

%% file: sections/5_experiments.tex
\section{Experiments}
\label{sec:5_experiments}

\begin{figure*}[!t]
\vspace*{2mm}
\centering
\includegraphics[width=0.97\textwidth]{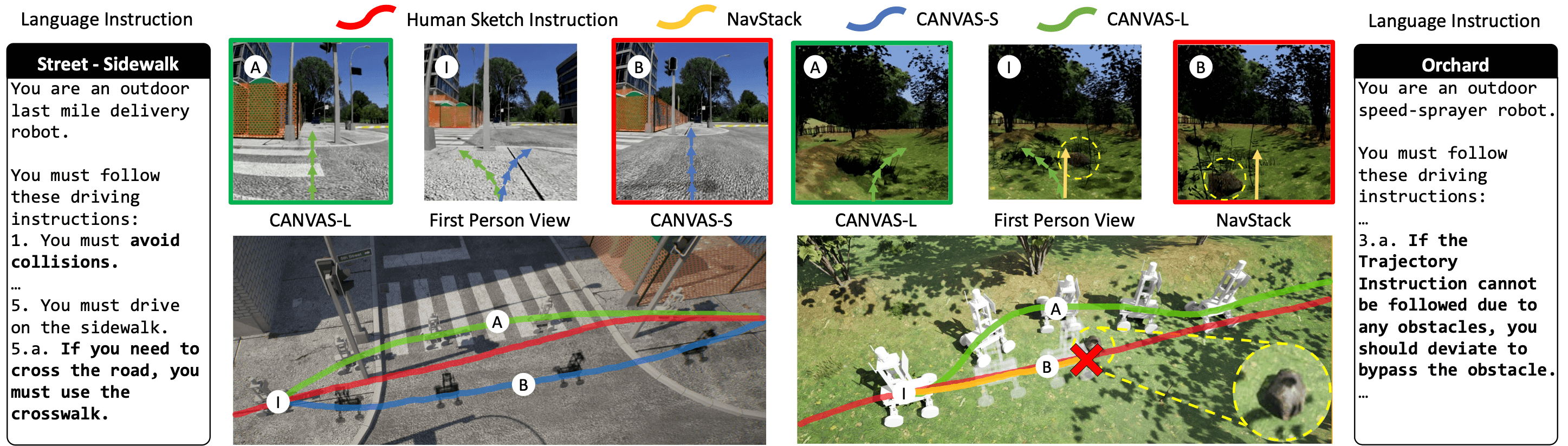}
\caption{The left side of the figure compares \mymethod-L and \mymethod-S, showing \mymethod-L using the crosswalk despite a misleading sketch instruction. The right side compares \mymethod-L and \mybaseline, illustrating \mymethod-L avoiding small obstacles, such as rocks.}
\label{fig:4_success_case_study}
\vspace*{-4mm}
\end{figure*}

In this section, we aim to answer the following questions:

\begin{enumerate}
    \setlength{\itemsep}{0.1em}
    \item [A.] Can \mymethod handle a variety of commonsense-aware navigation tasks in simulated environments? 
    \item [B.] Can \mymethod be transferred to a real-world environment in a zero-shot manner?
    \item [C.] How much does leveraging the pre-trained weights of the VLM enhance \mymethod’s performance?
    \item [D.] When \mybaseline fails, in what ways can \mymethod succeed?
    \item [E.]  Is \mymethod fast enough for real-time navigation?
\end{enumerate}

\textbf{Experimental Setup.}
In COMMAND, successful navigation requires the robot to reach the target location without collisions, while also respecting commonsense constraints.
As a result, four key metrics were used to evaluate performance: SR, CR, TDD, and IVR.
Success Rate (SR) represents the proportion of successful episodes, while Collision Rate (CR) captures the proportion of episodes with collisions.
Trajectory Deviation Distance (TDD) measures how closely the model follows human demonstrations, using the interquartile mean of \textit{Frechet distances}.
Finally, Instruction Violation Rate (IVR) assesses the proportion of episodes where human evaluators identified violations of commonsense constraints, such as keeping to the right lane or using crosswalks.
TDD was calculated only for success cases, as including failure cases would skew the metric.

We compare \mymethod with \textbf{\mybaselinelong}~\cite{rosnavstack}, a straightforward yet effective rule-based navigation system.
For this system, we converted the sketch instructions into step-by-step, point-to-point inputs, but language instructions could not be accommodated.
The same hyperparameters were used for all experiments with \mybaseline.
We evaluate two variations of \mymethod.
\textbf{\mymethod-S} modifies the original Idefics2 8B~\cite{idefics2} by swapping the vision encoder for SigLIP-L~\cite{siglip} and the text encoder for Qwen2-0.5B~\cite{yang2024qwen2}, reducing the model size from 8B to 0.7B to better accommodate real-world deployment.
In contrast, \textbf{\mymethod-L} retains the original Idefics2 8B~\cite{idefics2} architecture with its pre-trained weights. Both models were trained using LoRA~\cite{lora} with $r$=256, $a$=512, and a dropout rate of 0.1. We use the AdamW optimizer~\cite{adamw} with a different learning rates of 2e-5 for the LLM and 5e-5 for the projector and vision encoder, a batch size of 32, and 5 training epochs. Each model utilizes 128 waypoint tokens. In the main experiments, \mymethod models were inferred using a single NVIDIA H100 GPU. All experiments were evaluated over three iterations for each test dataset, with randomized starting orientations. In the real-world environment, SLAM with FAST-LIO2~\cite{fast-lio2} was used to find the robot’s current position.

\input{tables/3_1_result_sim}
\input{tables/3_2_result_sim_ivr}

\subsection{Results in the Simulated Environments}
\label{RQ1}

\myparagraph{Seen Environments}.
Table~\ref{tab:3_1_result_sim} shows \mymethod's performance in three seen environments: office, street (road, sidewalk), and orchard. 
Under precise instructions, \mymethod achieves similar SR and CR to \mybaseline in the office and street (road), where navigation is easier, indicating that \mymethod can learn the essential navigation behaviors effectively from human demonstrations. 
However, in more challenging environments like the street (sidewalk) and orchard, \mymethod significantly outperforms \mybaseline. A detailed analysis of \mymethod's performance is provided in Section \ref{RQ4}.

Table ~\ref{tab:3_2_result_sim_ivr} compares the IVR between \mymethod and \mybaseline.
In the street (road), commonsense constraints include lane adherence, while in the street (sidewalk), they involve crossing roads correctly and staying on the sidewalk.
Unlike NavStack's rule-based approach, \mymethod learns commonsense driving rules from human demonstrations, consistently achieves lower IVR even with misleading instructions.

\myparagraph{Unseen Environment}.
We exclude the gallery environment during training to evaluate \mymethod's performance in unseen settings. 
As demonstrated in Table~\ref{tab:3_1_result_sim}, \mymethod continues to show strong navigation capabilities, even in scenarios with noisy guidance.

\subsection{Results in the Real-World Environment}
\label{RQ2}

\begin{figure}[t!]
\vspace*{2mm}
\centering
\includegraphics[width=0.35\textwidth]{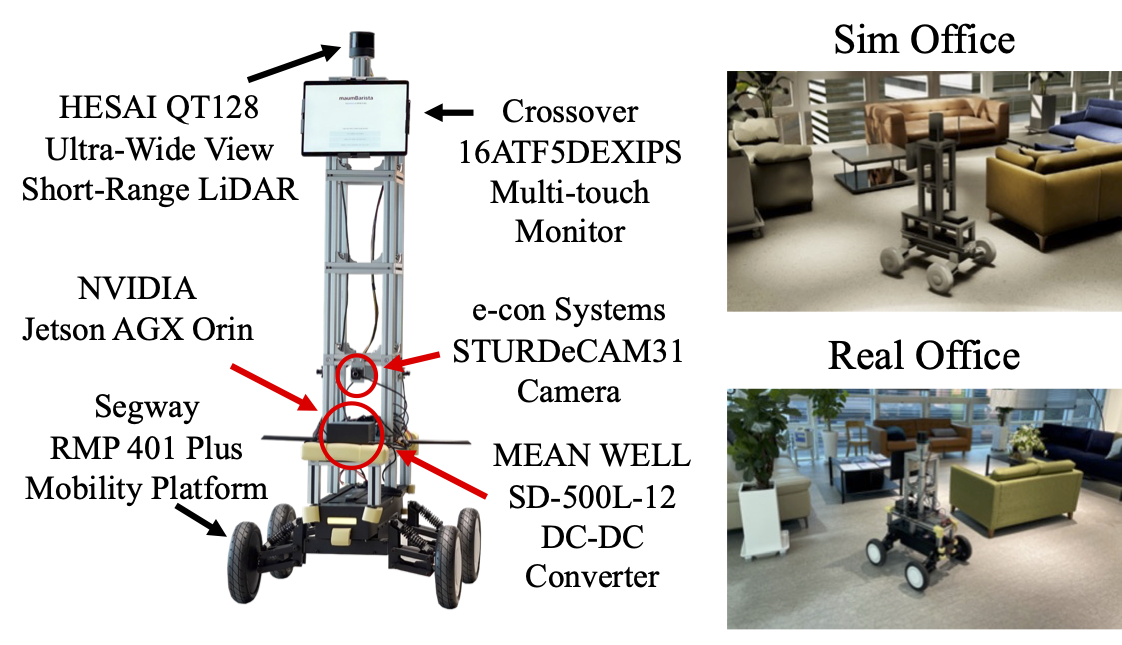}
\caption{
We developed a physical robot and created a realistic simulated environment that replicates real-world conditions.}
\label{fig:5_sim2real}
\vspace*{-3mm}
\end{figure}

While COMMAND collects human demonstrations across a variety of simulated environments designed to resemble real-world conditions, a potential concern is whether these simulations fully capture the complexity of the real world.
Therefore, it is important to demonstrate that the \mymethod's effective navigation in simulation can extend to real-world environments.
As shown in Figure~\ref{fig:5_sim2real}, to assess its real-world performance, we tested \mymethod in an actual office environment that was used as the basis for the simulation.
Despite being trained solely on simulated data, \mymethod demonstrated strong Sim2Real transfer capabilities, performing reliably in real-world scenario.

\input{tables/4_result_real}

\subsection{Ablation Study}
\label{RQ3}

We explore the importance of leveraging pre-trained weights from the VLM.
As demonstrated in Table~\ref{tab:5_ablation_study_pretrained_weight}, these weights were crucial for \mymethod's performance, especially in both unseen simulated and real-world settings. 
This indicates that the knowledge encapsulated in the pre-trained VLMs offered a strong foundation for \mymethod to learn how to incorporate them in developing a generalizable understanding of driving dynamics.

\input{tables/5_ablation_study_pretrained_weight}

\subsection{Additional Case Study}
\label{RQ4}

\begin{figure}[t!]
\vspace*{2mm}
\centering
\includegraphics[width=0.48\textwidth]{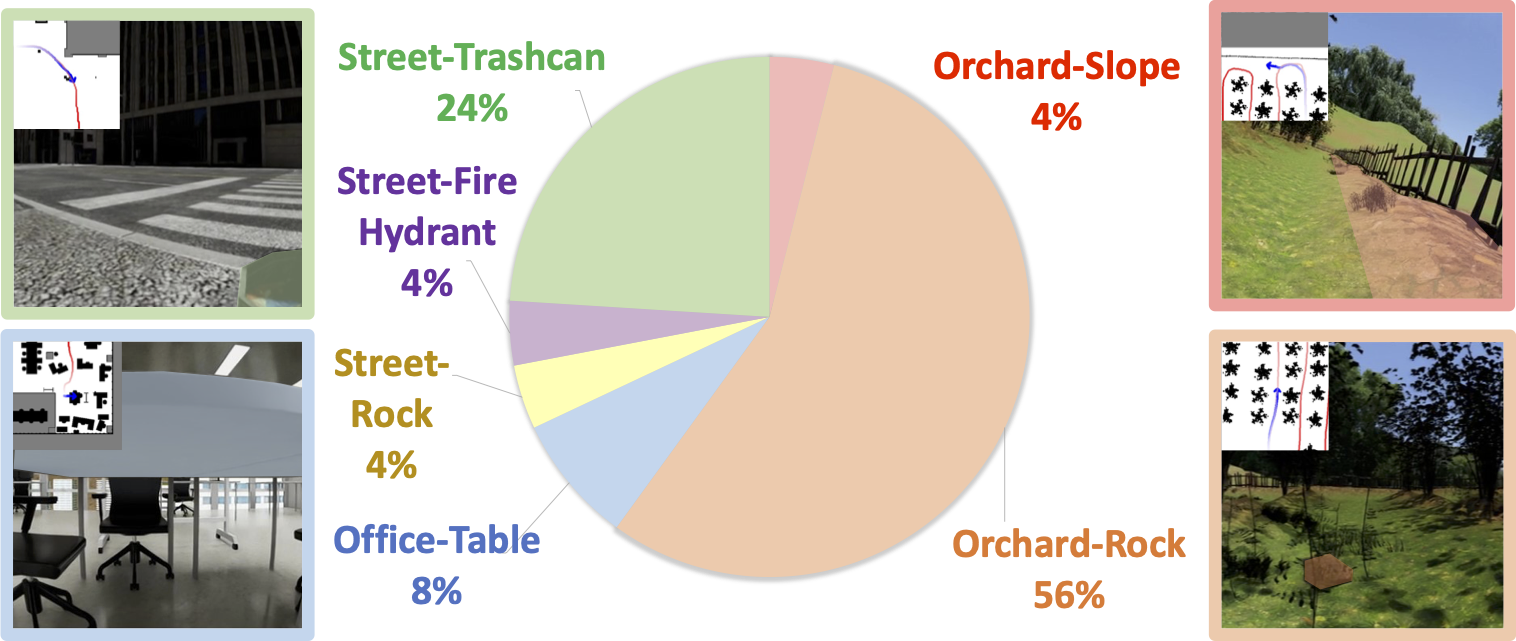}
\caption{We classify the failure cases of \mybaseline~\cite{rosnavstack} in various simulated environments.}
\label{fig:6_failure_case_study}
\vspace*{-3mm}
\end{figure}

We perform a qualitative analysis to examine the factors behind the success of \mymethod compared to \mybaseline~\cite{rosnavstack}.
Figure~\ref{fig:6_failure_case_study} highlights typical failure cases for \mybaseline, where the robot is unable to reach its destination.
In 56\% of these cases, failures result from stumbling over rocks in the orchard environment. Figure~\ref{fig:4_success_case_study} compares \mymethod and \mybaseline in this primary failure scenario.
The orchard has uneven terrain, and \mybaseline struggles to avoid small but hazardous obstacles like rocks because its limited perception cannot distinguish rocks from passable areas such as grass. 

Tuning the hyperparameters of \mybaseline exacerbates the issue: setting the obstacle's height threshold lower causes the system to misclassify grass as an obstacle, and the robot LiDAR sensor, which suffers from significant vibration, struggles to navigate with a small threshold.
In contrast, \mymethod utilizes visual inputs from the camera to reliably detect unexpected obstacles and assesses their navigation risk based on learned experiences from demonstrations.

\subsection{Real-Time Navigation Feasibility Study}
\label{RQ5}

Finally, we evaluate the feasibility of deploying \mymethod in real-time applications. The NVIDIA AGX Orin, integrated into our robot, demonstrates on-device inference capabilities with an average latency of 400ms for \mymethod-S and 800ms for \mymethod-L, highlighting its potential for efficient real-world navigation tasks without substantial delays.

%% file: tables/3_1_result_sim.tex
\begin{table}[h]
\centering
\vspace*{2mm}
\setlength{\tabcolsep}{3.5pt}
\resizebox{0.95\linewidth}{!}{
    \begin{tabular}{c|ccc|ccc|c}
        \hline
        \multirow{2}{*}{Method} & \multicolumn{3}{c|}{Precise} & \multicolumn{3}{c|}{Misleading} & Total\\
        & SR($\uparrow$) & CR($\downarrow$) & TDD($\downarrow$) & SR($\uparrow$) & CR($\downarrow$) & TDD($\downarrow$) & SR($\uparrow$)\\
        \hline
        \multicolumn{8}{c}{Seen Environment}\\
        \hline
        \multicolumn{8}{c}{\textit{Office}}\\
        \mybaseline & 87\% & 13\% & 0.846m & 0\%* & 100\%* & - & - \\
        \mymethod-S & \textbf{100}\% & \textbf{0}\% & \textbf{0.730m} & 87\% & 13\% & 0.843m & 93\% \\
        \ours \mymethod-L & \textbf{100}\% & \textbf{0}\% & 0.802m & \textbf{100}\% & \textbf{0}\% & \textbf{0.753m} & \textbf{100\%} \\
        \hdashline
        \multicolumn{8}{c}{\textit{Street (Road)}}\\
        \mybaseline & \textbf{100}\% & \textbf{0}\% & 1.654m & 0\%* & 100\%* & - & -\\
        \mymethod-S & \textbf{100}\% & \textbf{0}\% & 1.189m & \textbf{100}\% & \textbf{0}\% & \textbf{1.075m} & \textbf{100\%} \\
        \ours \mymethod-L & 97\% & 3\% & \textbf{1.117m} & 97\% & 3\% & 1.236m & 97\% \\
        \hdashline
        \multicolumn{8}{c}{\textit{Street (Sidewalk)}}\\
        \mybaseline & 53\% & 53\% & 1.450m & 0\%* & 100\%* & - & - \\
        \mymethod-S & 60\% & 40\% & 1.451m & 47\% & 53\% & 2.379m & 54\% \\
        \ours \mymethod-L & \textbf{87}\% & \textbf{13}\% & \textbf{1.394m} & \textbf{53}\% & \textbf{47}\% & \textbf{1.839}m & \textbf{70\%} \\
        \hdashline
        \multicolumn{8}{c}{\textit{Orchard}}\\
        \mybaseline & 0\% & 87\% & - & 0\%* & 100\%* & - & - \\
        \mymethod-S & \textbf{73}\% & 60\% & \textbf{1.561m} & \textbf{60}\% & \textbf{33}\% & 1.448m & \textbf{67\%} \\
        \ours \mymethod-L & 67\% & \textbf{47}\% & 1.759m & \textbf{60}\% & 53\% & \textbf{1.392m} & 64\% \\
        \hline
        \multicolumn{8}{c}{Unseen Environment}\\
        \hline
        \multicolumn{8}{c}{\textit{Gallery}}\\
        \mybaseline & \textbf{100}\% & \textbf{0}\% & 0.783m & 0\%* & 100\%* & - & - \\
        \mymethod-S & 87\% & 13\% & \textbf{0.773m} & \textbf{33}\% & \textbf{66}\% & 0.938m & 60\%\\
        \ours \mymethod-L & \textbf{100}\% & 7\% & 0.9m  & \textbf{33}\% & \textbf{66}\% & \textbf{0.856m} & \textbf{67\%} \\
        \hline
    \end{tabular}
}
\caption{Evaluation results on simulated environments. *: NavStack was not tested in the misleading scenario because it is not equipped to handle such situations.}

\label{tab:3_1_result_sim}
\vspace*{-2.5mm}
\end{table}

%% file: tables/3_2_result_sim_ivr.tex
\begin{table}[h]
\centering
\scriptsize
\begin{tabular}{c|cccc}
    \hline
    \multirow{2}{*}{Environment} & \multirow{2}{*}{Method} & Precise & Misleading \\
    & & IVR($\downarrow$) & IVR($\downarrow$) \\
    \hline
    \multirow{3}{*}{Street (Road)} & NavStack & 7\% & 100\%* \\
    &\mymethod-S & \textbf{0\%} & \textbf{7\%} \\
    & \multicolumn{1}{>{\columncolor{LimeGreen!30}}c}{\mymethod-L} & \multicolumn{1}{>{\columncolor{LimeGreen!30}}c}{17\%} & \multicolumn{1}{>{\columncolor{LimeGreen!30}}c}{30\%} \\
    \hline
    \multirow{3}{*}{Street (Sidewalk)} & NavStack & 7\% & 100\%* \\
    & \mymethod-S & \textbf{0\%} & 26\% \\
    & \multicolumn{1}{>{\columncolor{LimeGreen!30}}c}{\mymethod-L} & \multicolumn{1}{>{\columncolor{LimeGreen!30}}c}{\textbf{0\%}} & \multicolumn{1}{>{\columncolor{LimeGreen!30}}c}{\textbf{13\%}} \\
    \hline
\end{tabular}
\caption{Evaluation of violation rates for commonsense constraints in a street environment.}
\label{tab:3_2_result_sim_ivr}
\vspace*{-4.5mm}
\end{table}

%% file: tables/4_result_real.tex
\begin{table}[h]
\centering
\vspace*{-2mm}
\begin{tabular}{l|ccc}
    \hline
    \multirow{2}{*}{Method} & Precise & Misleading & Total \\
    & SR($\uparrow$) & SR($\uparrow$) & SR($\uparrow$)\\
    \hline
    \mybaseline & \textbf{100\%} & 0\%* & - \\
    \hdashline
    \mymethod-S & 77\% & \textbf{60\%} & \textbf{69\%} \\
    \ours \mymethod-L & 93\% & 33\% & 63\% \\
    \hline
\end{tabular}
\caption{Evaluation results on real environments.}
\label{tab:4_result_real}
\vspace*{-4mm}
\end{table}

%% file: tables/5_ablation_study_pretrained_weight.tex
\begin{table}[h]
\vspace*{-2mm}
\centering
\resizebox{0.95\linewidth}{!}{%
    \begin{tabular}{l|cccc}
        \hline
        \multirow{2}{*}{Environment} & \multirow{2}{*}{Method} & Precise & Misleading & Total \\
        & & SR($\uparrow$) & SR($\uparrow$) & SR($\uparrow$) \\
        \hline
        \multirow{2}{*}{Seen - Office} & \multicolumn{1}{>{\columncolor{LimeGreen!30}}c}{\mymethod-L} & \multicolumn{1}{>{\columncolor{LimeGreen!30}}c}{\textbf{100\%}} & \multicolumn{1}{>{\columncolor{LimeGreen!30}}c}{\textbf{100\%}} & \multicolumn{1}{>{\columncolor{LimeGreen!30}}c}{\textbf{100\%}} \\
        & w/o Pre-training & \textbf{100\%} & 87\% & 93\% \\
        \hline
        \multirow{2}{*}{Unseen - Gallery} & \multicolumn{1}{>{\columncolor{LimeGreen!30}}c}{\mymethod-L} & \multicolumn{1}{>{\columncolor{LimeGreen!30}}c}{\textbf{100\%}} & \multicolumn{1}{>{\columncolor{LimeGreen!30}}c}{33\%} & \multicolumn{1}{>{\columncolor{LimeGreen!30}}c}{\textbf{67\%}} \\
        & w/o Pre-training & 60\% & \textbf{40\%} & 50\% \\
        \hline
        \multirow{2}{*}{Real - Office} & \multicolumn{1}{>{\columncolor{LimeGreen!30}}c}{\mymethod-L} & \multicolumn{1}{>{\columncolor{LimeGreen!30}}c}{\textbf{93\%}} & \multicolumn{1}{>{\columncolor{LimeGreen!30}}c}{\textbf{33\%}} & \multicolumn{1}{>{\columncolor{LimeGreen!30}}c}{\textbf{63\%}} \\
        & w/o Pre-training & 73\% & \textbf{33\%} & 53\% \\
        \hline
    \end{tabular}
}
\caption{Ablation study on the effect of VLM pre-training.}
\label{tab:5_ablation_study_pretrained_weight}
\vspace*{-4mm}
\end{table}

%% file: sections/6_conclusion.tex
\section{Conclusion}
\label{sec:6_conclusion}

We present \mymethod, a novel commonsense-aware navigation system that learns from human demonstrations through imitation learning. 
\mymethod allows intuitive human instructions using abstract sketches and natural language while leveraging commonsense reasoning to bridge the gap between vague human guidance and concrete robot actions.
With the COMMAND dataset for imitation learning and pre-trained vision-language models, \mymethod allows robots to understand implicit human intent and make decisions aligned with human expectations.
Experiments show that \mymethod outperforms \mybaselinelong, a strong rule-based system, with higher success rates, fewer collisions, and better trajectory alignment with human demonstrations, all while adhering to commonsense constraints.
Additionally, \mymethod exhibits strong performance in both unseen and real-world environments, highlighting its generalization capabilities.
By open-sourcing the COMMAND dataset and \mymethod, we hope to contribute to active research on imitation learning techniques for commonsense reasoning in robot navigation.

%% file: sections/acknowledgment.tex
\section*{Acknowledgments}

This research was partly supported by GINT.
The authors would like to thank GINT for providing feedback on the agricultural spraying robot.